\documentclass[journal]{IEEEtran}

\usepackage{subfig}
\usepackage[pdftex]{graphicx}

\ifCLASSINFOpdf
\else
\fi
\usepackage{amsmath}
\usepackage{amsfonts}
\usepackage{multirow}
\usepackage{comment}

\usepackage{algpseudocode}

\usepackage{algorithm}

\usepackage{array}

\usepackage{times}
\usepackage{epsfig}
\usepackage{graphicx}
\usepackage{amssymb}
\usepackage{xcolor}
\usepackage{makecell}
\usepackage{balance}

\begin{document}
%
\title{Attention-based Image Upsampling}
%
%
%

\author{Souvik Kundu*,
        Hesham Mostafa*,
        Sharath Nittur Sridhar*,
        Sairam Sundaresan*
\IEEEcompsocitemizethanks{\IEEEcompsocthanksitem S. Kundu is a Ph.D. student at the University of Southern California, USA. The work was done during his internship at Intel Labs. H. Mostafa, S. N. Sridhar, and S. Sundaresan are with Intel Labs, San Diego, USA.
                E-mail: souvikku@usc.edu, \{hesham.mostafa, sharath.nittur.sridhar, sairam.sundaresan\}@intel.com. 
                }
\thanks{* Equal contributions. Authors listed in alphabetical order.}}

%
%

\markboth{Attention-based Image Upsampling}%
{Shell \MakeLowercase{\textit{et al.}}: Bare Demo of IEEEtran.cls for IEEE Journals}
%



\maketitle

\begin{abstract}
  Convolutional layers are an integral part of many deep neural network solutions in computer vision. Recent work shows that replacing the standard convolution operation with mechanisms based on self-attention leads to improved performance on image classification and object detection tasks. In this work, we show how attention mechanisms can be used to replace another canonical operation: strided transposed convolution. We term our novel attention-based operation attention-based upsampling since it increases/upsamples the spatial dimensions of the feature maps. Through experiments on single image super-resolution and joint-image upsampling tasks, we show that attention-based upsampling consistently outperforms traditional upsampling methods based on strided transposed convolution or based on adaptive filters while using fewer parameters. We show that the inherent flexibility of the attention mechanism, which allows it to use separate sources for calculating the attention coefficients and the attention targets, makes attention-based upsampling a natural choice when fusing information from multiple image modalities. 
\end{abstract}


%
\IEEEpeerreviewmaketitle

\section{Introduction}
%
%
%
%
\IEEEPARstart{C}{onvolutional} layers are an integral part of most deep neural networks used in vision applications such as image recognition \cite{krizhevsky2012imagenet, simonyan2014very, szegedy2015going}, image segmentation \cite{sermanet2013overfeat, redmon2017yolo9000}, joint image filtering \cite{li2019joint, kim2020deformable}, image super-resolution \cite{5466111, 937655, dong2016accelerating, bevilacqua2012low, zeyde2010single} to name a few. The convolution operation is translationally equivariant, which when coupled with pooling, introduces useful translation invariance properties in image classification tasks. Computationally, local receptive fields and weight sharing are highly beneficial in reducing the number of floating point operations (FLOPs) and the parameter count, respectively, compared to fully connected layers. However, using the same convolutional weights at all pixel positions might not necessarily be the optimal choice. Motivated by the success of attention mechanisms \cite{chorowski2015attention, vaswani2017attention} in natural language processing tasks, several recent methods replace the convolution operation with a self-attention mechanism \cite{ramachandran2019stand, bello2019attention}. While the receptive field size does not necessarily change, the use of self-attention allows the pixel aggregation weights (attention coefficients) to be content-dependent. This, in effect, results in an adaptive kernel whose weights vary across the spatial locations based on the receptive field contents. 

In this paper, we focus on another fundamental operation in vision applications which has received relatively little attention: strided transposed convolution or deconvolution. Unlike the standard strided convolution operation which reduces the input's spatial dimension by the stride factor, strided transposed convolution {\it increases} the input's spatial dimension by the stride factor. Strided transposed convolution is thus a natural operation to use in image upsampling tasks. We replace the fixed kernel used in the strided transposed convolution by a self-attention mechanism that calculates the kernel weights at each pixel position based on the contents of the receptive field.


We test our new attention-based upsampling mechanism on two popular tasks: single image super-resolution and joint-image upsampling. We show on various benchmarks in the single-image super-resolution task that our attention-based upsampling mechanism performs consistently better than using standard strided transposed convolution operations. Our attention-based mechanism also uses considerably fewer parameters. Thus, the proposed attention-based upsampling layer can be used as as a drop-in replacement of the strided transposed convolution layer, which improves model performance due to the content-adaptive nature of the attention-generated kernels.

In joint-image upsampling tasks, the goal is to upsample an image from one modality using a high-resolution image from another modality. Unlike the standard strided transposed convolution operation, our attention-based formulation produces three intermediate tensors, namely the query, key, and value tensors. This allows us more flexibility in defining the attention-based upsampling mechanism as we do not necessarily have to obtain these tensors from the same set of feature maps. Indeed, we show that in joint-image upsampling tasks that involve multiple input modalities (for example a depth map and an RGB image), obtaining the query and key tensors from one modality and the value tensor from another modality naturally  allows our attention-based upsampling mechanism to combine information from multiple modalities. The conventional strided transposed convolution operation does not have this ability. We thus compare against joint image upsampling models of similar complexity and show that our attention-based joint upsampling model is more accurate and parameter-efficient.
 


The reminder of this paper is arranged as follows. Section \ref{sec:related} describes related work. Details of our proposed method and experimental evaluations are presented in Section \ref{sec:method} and \ref{sec:expt_res}, respectively. We provide our conclusions in Section \ref{sec:concl}.

\section{Related Work}
\label{sec:related}
\subsection{Self-Attention}

Originally introduced in \cite{bahdanau2014neural}, attention mechanisms have displayed impressive performance in
sequence to sequence modeling tasks, where they significantly outperform traditional methods based on recurrent neural networks~\cite{Vaswani_etal17}. An attention mechanism aggregates features from multiple inputs where the aggregation weights are dynamically calculated based on the input features. Attention mechanisms differ in the way they calculate the aggregation weights (attention coefficients)~\cite{Vaswani_etal17,Luong_etal15,Graves_etal14}. In this paper, we primarily use the scaled dot product attention mechanism of ~\cite{Vaswani_etal17}, which is an extension of the unscaled dot product method of \cite{Luong_etal15}. 

Beyond their use in sequence to sequence tasks, attention mechanisms have more recently been used to replace or augment convolutional neural networks (CNNs)~\cite{Ramachandran_etal19,Bello_etal19,Zhao_etal20}. Instead of using a fixed kernel to aggregate features from all pixels within a receptive field, attention-based convolution uses an attention mechanism to dynamically calculate these aggregation weights based on the contents of the receptive field. This offers more flexibility as the kernel is now content-dependent and not fixed. This leads to improved accuracy on several classification and object detection tasks~\cite{Ramachandran_etal19,Bello_etal19,Zhao_etal20}. It is improtant to note that these convolutional attention models are fundamentally different from previous attention based vision models that use attention to preferentially aggregate information from parts of the input image~\cite{Xu_etal15,Mnih_etal14}, but do not modify the convolution operation itself.

\subsection{Joint Image Upsampling}
In joint image upsampling tasks, in addition to the low-resolution image, the network also receives a corresponding high-resolution image, but from a different modality. Information from the high-resolution modality (the guide modality) can be used to guide the upsampling of the low-resolution modality (the target modality). In particular, image filtering with the help of a guidance image has been used on a variety of tasks which include depth map enhancement \cite{yang2007spatial, ferstl2013image, ham2017robust}, cross-modality image restoration \cite{shen2015mutual}, texture removal \cite{zhang2014rolling}, dense correspondence \cite{hosni2012fast}, and  semantic segmentation \cite{barron2016fast}. The basic idea of joint image filtering is to transfer the structural details from a high-resolution guidance image to a low-resolution target through estimation of spatially-varying kernels from the guidance. Classical approaches of joint image filtering  focus on hand-crafted kernels \cite{he2010guided}.
For example, both the bilateral filter \cite{tomasi1998bilateral} and the guided filter \cite{he2012guided} leverage the local structure of the guidance image using spatially-variant Gaussian kernels and matting Laplacian kernels \cite{levin2007closed}, respectively, without handling  inconsistent structures in the guidance and target
images. The SD filter \cite{ham2017robust} constructs spatially-variant kernels from both guidance and target images and exploits common structures by formulating joint image filtering as an optimization problem. Recent advances in learnable upsampling mechanisms using convolutional layers achieve state-of-the-art performance at various upsampling factors \cite{kim2020deformable, li2016deep}. In particular, \cite{li2016deep} proposes a deep joint filtering (DJF) CNN based model to learn the spatially-variant kernels by non-linearly mixing activation values of spatially-invariant kernels. On the the other hand,  the fast deformable kernel network (FDKN) \cite{kim2020deformable} exploits the advantage of spatially-variant kernels by learning to generate kernels at each spatial positions, together with learning the receptive field structure at each position, outperforming previous approaches. The closest technique to our method uses the feature maps from the guide modality to adaptively modify the convolutional kernels applied to the target modality image~\cite{Su_etal19}. Similar to our approach, these kernel adaptations are different at each position. 

\subsection{Single Image Super Resolution}
Increasing the resolution of images, or image upscaling is a common task with many variations in computer vision. Initial approaches were based on non-learnable interpolation methods such as bilinear or bicubic interpolation~\cite{Keys_etal81}. Non-learnable interpolation methods, however, significantly under-perform learning based methods that utilize a deep network trained in a supervised setting to produce the high-resolution image from the low-resolution input~\cite{Wang_etal20}. These two are often combined where a non-learnable interpolation mechanism is used to upsample the image, followed by a learnable CNN to produce the final high-resolution output~\cite{Dong_etal15}, or combined in the reverse order where a CNN is first applied to the low-resolution image followed by non-learnable interpolation to upsample the final image~\cite{Dong_etal16}. The upsampling mechanism itself can be learned; this is the case when using the strided transposed convolution, or the deconvolution~\cite{Dumoulin_etal16} operation to upsample the image. An alternative learned upsampling mechanism is sub-pixel convolution~\cite{Shi_etal16}, where $F*S^2$ feature maps of dimensions $H\times W$ are reshaped into $F$ feature maps of dimensions $H*S \times W*S$, where $S$ is the upsampling factor. 

\section{Methods}
\label{sec:method}

\begin{figure}[!t]
\centering
\includegraphics[width=2.0in]{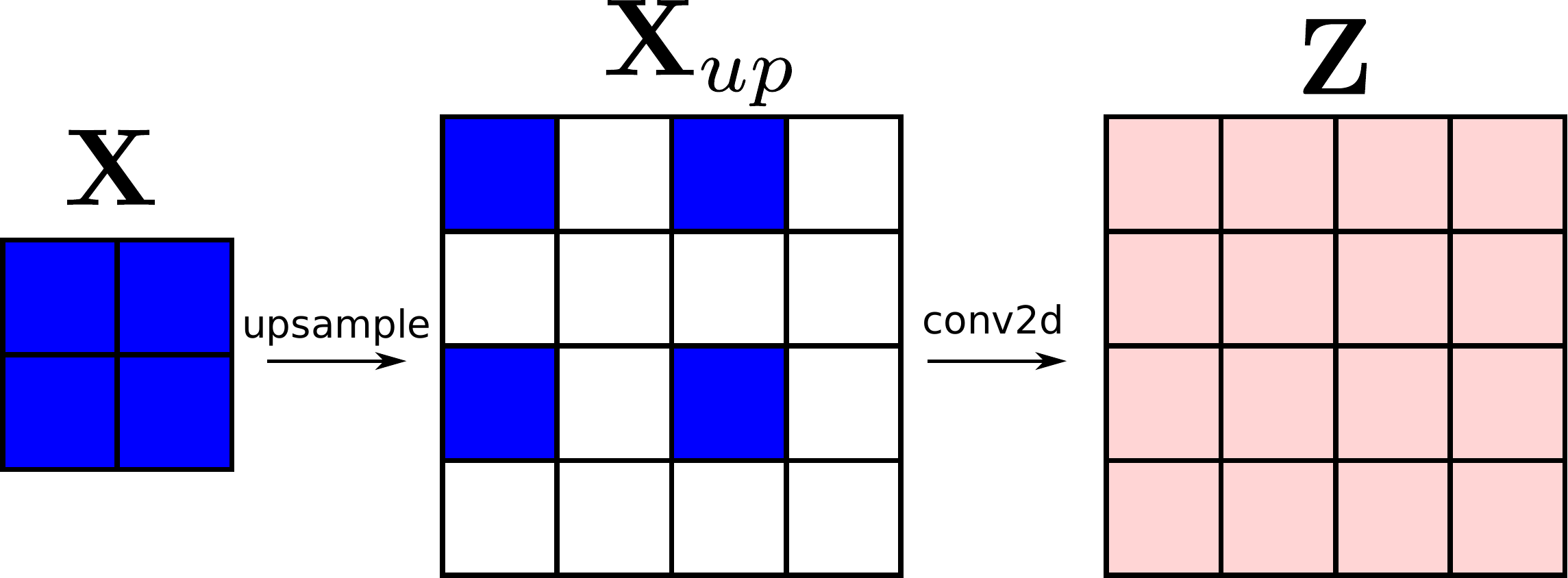}
\caption{Illustration of the strided transposed convolution with stride 2.}
\label{fig:stc}
\end{figure}

\subsection{Background}
\subsubsection{Scaled dot product attention}
The attention mechanism we use is a scaled dot product attention mechanism similar to that proposed in ref.~\cite{Vaswani_etal17}, which we summarize here. The inputs to the attention mechanism are query, key, and value tensors: ${\bf Q}\in R^{N\times F_K}$, ${\bf K}\in \mathbb{R}^{N \times F_K}$, and ${\bf V}\in \mathbb{R}^{N\times F_V}$, respectively. $N$ is the number of input and output tokens. $F_K$ is the dimension of the key and query vectors for each token, and $F_V$ is the dimension of the value vector at each token. The output of the attention mechanism, ${\bf Y}\in \mathbb{R}^{N\times F_V}$ is given by:
\begin{align}
{\bf A} & = \frac{{\bf Q}{\bf K}^T}{\sqrt{F_K}}, \label{eq:attention_nopos}\\
{\bf Y} & = softmax({\bf A}){\bf V},
\end{align}
where ${\bf A}\in \mathbb{R}^{N \times N}$ is the matrix of attention coefficients and the softmax operation is applied row-wise. 

Equation~\ref{eq:attention_nopos} is invariant to the ordering of the input tokens, which is undesirable as the order contains important information. A common approach to address this issue is to add positional encoding vectors to the query and key vectors to inject positional information. These positional encodings can either be absolute~\cite{Vaswani_etal17}, i.e, they encode the absolute position of the token's query and key vectors, or they can be relative, i.e, they encode the relative position between the query/key vectors at one position and the query/key vectors at a second position. We primarily focus on relative positional encoding methods inspired by ref.~\cite{Shaw_etal18}. With relative positional encoding, instead of Eq.~\ref{eq:attention_nopos}, the entries of the attention matrix ${\bf A}$ are given by:
\begin{equation}
a_{ij} = \frac{{\bf q}_i^T ({\bf k_j} + {\bf p}_{i-j})}{\sqrt{F_K}},
\end{equation}
where ${\bf p}_{i-j}$ is the relative positional encoding term, ${\bf q}_i$ is the query vector at output position $i$, and ${\bf k}_j$ is the key vector at input position $j$. 

\subsubsection{Strided transposed  convolution}
Convolution is one of the prototypical operations in computer vision. In this paper, we are primarily interested in a variant of the convolution operation which is the strided transposed convolution. This operation is also called fractionally strided convolution or deconvolution~\cite{Dumoulin_etal16}. We use these terms interchangeably throughout the paper. The architectural parameters of a strided transposed convolution is the kernel size($K$), the stride($S$), the number of input feature maps($C_{in}$), and number of output feature maps($C_{out}$). We do not consider the padding parameters, because we always choose them so that the operation exactly scales all spatial dimensions by a factor of $S$. Given an input set of feature maps ${\bf X} \in \mathbb{R}^{C_{in}\times H \times W}$ where $H$ and $W$ are are the height and width of the input feature maps, the strided transposed convolution operation first zero-upsamples ${\bf X}$ by a factor $S$ to yield ${\bf X}_{up} = zero\_upsample({\bf X};S) \in \mathbb{R}^{C_{in}\times S*H \times S*W} $, where $S-1$ zeros are inserted between adjacent entries, and after the final row and column:
\vspace{2mm}
\begin{equation}
  X_{up}[c,i,j] = \begin{cases}
    X[c,i//S,j//S], & \parbox{3cm}{if ($i$ mod $S=0$) and ($j$ mod $S=0$)} \\
    0, & \text{otherwise},
  \end{cases}
  \label{eq:zero_upsample}
\end{equation}
\vspace{1mm}
\noindent

\begin{figure}[!t]
\centering
\includegraphics[width=3in]{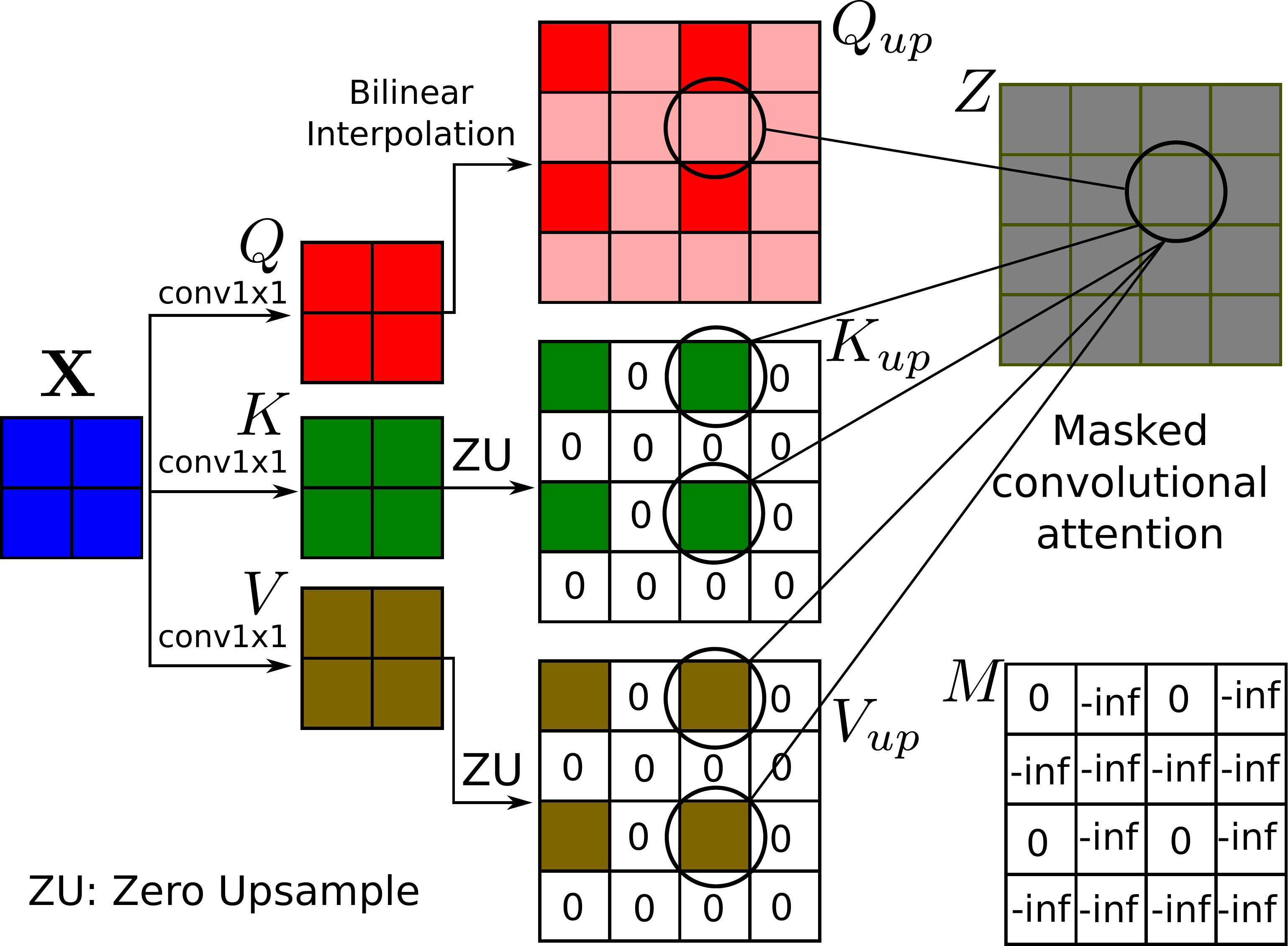}
\caption{The attention-based upsampling operation for $S=2$ and $K=3$. For a particular output position, position $(1,2)$, we show its dependence on entries in ${\bf Q}_{up}$, ${\bf K}_{up}$, and ${\bf V}_{up}$. To produce the output at position $(1,2)$, we use a masked attention mechanism in a $K \times K$ window around (1,2). There are only two input positions allowed by the mask (position $(0,2)$ and position $(2,2)$). We use the query vector at position (1,2) to obtain the normalized attention coefficients through a dot product with the key vectors at these two positions followed by a softmax. The resulting two normalized attention coefficients are then used to sum the value vectors at the two allowed positions to yield the output vector.}
\label{fig:attention_upsampling}
\end{figure}

where $//$ denotes integer division and indexing is 0-based. ${\bf X}_{up}$ is then fed to a standard convolution operation with stride one, kernel size $K$, and padding $(K-1)//2$ (we always assume $K$ is odd) to yield ${\bf Z}\in \mathbb{R}^{C_{out}\times S*H \times S*W}$:
\vspace{2mm}
\begin{equation}
{\bf Z} = conv2d({\bf X}_{up},{\bf W},padding = (K-1)//2),
\end{equation}
\vspace{2mm}
\noindent
where ${\bf W}\in \mathbb{R}^{C_{out} \times C_{in} \times K \times K}$ is the learnable kernel tensor. The strided transposed convolution operation is illustrated in Fig.~\ref{fig:stc}.

\subsubsection{Attention-based convolution}
Multiple prior work has looked into replacing the standard convolution operation by an attention mechanism~\cite{Ramachandran_etal19,Bello_etal19,Zhao_etal20}. In each receptive field window, the attention mechanism is responsible for generating a set of aggregation weights (attention coefficients) that are used by the center pixel to aggregate the features of the pixels in the receptive field window. Attention can thus be seen as a mechanism to generate a convolutional kernel at each spatial position. Unlike the convolutional kernel in standard CNNs which is shared across all spatial positions, the attention-generated kernels depend on the  contents of the receptive field and are thus different at each spatial position.

\begin{figure*}[!t]
\centering
\includegraphics[width=1.0\textwidth]{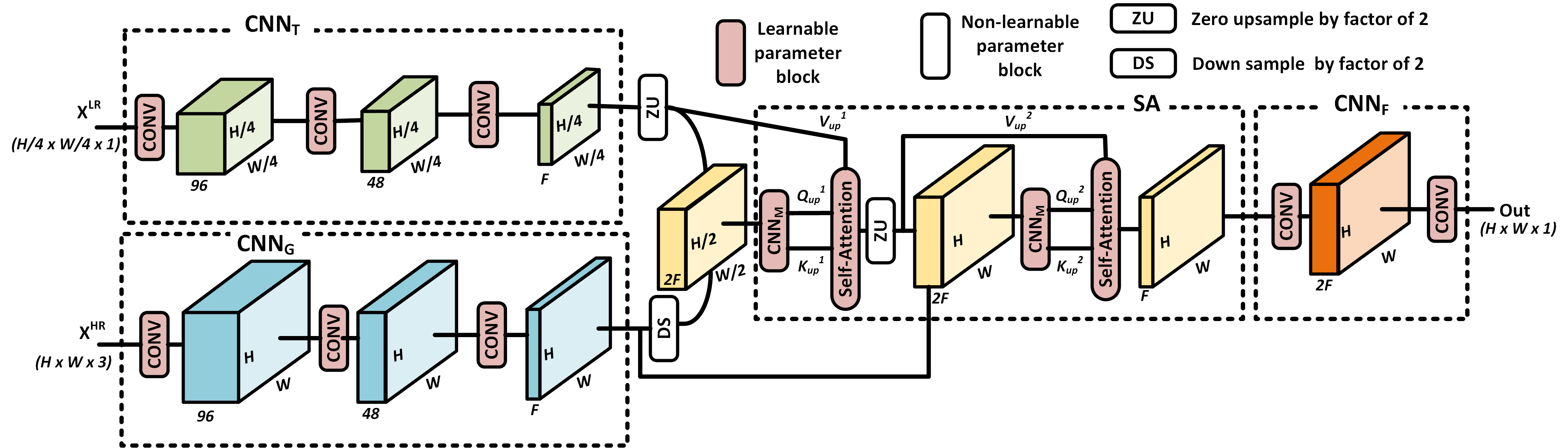}
\caption{Network architecture of the self-attention based deep joint upsampling model for an upsampling factor of 4. The proposed model consists of four major components. Similar to \cite{li2016deep} the sub-networks $CNN_T$ and $CNN_G$ aim to extract informative feature responses from the target and guidance images, respectively. The responses are then fed to a self-attention (SA) block to learn pixel specific contextual information, before being fed to the final convolutional block $CNN_F$. The desired reconstructed image is available at the output of $CNN_F$. $CNN_M$ takes the concatenated low-resolution and high-resolution features to produce the query and key tensors at each upsampling stage. All convolutional layers (except the final layer) and self-attention layers are followed by a ReLU non-linearity (not shown for simplicity).}
\label{fig:djf_localsa_block}
\end{figure*}

We are primarily interested in a dot product convolutional attention mechanism similar to that introduced in ref.~\cite{Ramachandran_etal19}. Given an input set of feature maps ${\bf X} \in \mathbb{R}^{C_{in}\times H \times W}$, three standard  $1\times 1$ convolution operations are used to obtain the query, key, and value tensors:
\begin{align}
\begin{split}
  {\bf Q} & = conv1d({\bf X},{\bf W}^Q),\\
  {\bf K} & = conv1d({\bf X},{\bf W}^K),\\
  {\bf V} & = conv1d({\bf X},{\bf W}^V),\label{eq:three_convs}
  \end{split}
\end{align}
where ${\bf Q},{\bf K},{\bf V} \in \mathbb{R}^{C_{out}\times H \times W}$. ${\bf W}^Q,{\bf W}^K,{\bf W}^V \in \mathbb{R}^{C_{out}\times C_{in} \times 1 \times 1}$ are the weight tensors of the three $1\times 1$ convolution operations. The output feature tensor ${\bf Z}\in \mathbb{R}^{C_{out}\times H \times W}$ is given by:
\begin{equation}
{\bf z}_{ij} = \sum\limits_{a,b \in \mathcal{N}_K(i,j)}softmax_{ab}\left({\bf q}_{ij}^T({\bf k}_{ab} + {\bf p}_{a-i,b-j})\right) {\bf v}_{ab},
\label{eq:vanilla_attn_conv}
\end{equation}
where $\mathcal{N}_K(i,j)$ is the $K\times K$ spatial neighborhood around position $(i,j)$. ${\bf z}_{ij},{\bf q}_{ij},{\bf k}_{ij},{\bf v}_{ij} \in \mathbb{R}^{C_{out}}$ are the output, query, key, and value feature vectors at position $(i,j)$, respectively. The softmax operator ensures that the attention coefficients within each $K\times K$ receptive field windows sum to one. The relative positional encoding vector ${\bf p}_{a-i,b-j}\in \mathbb{R}^{C_{out}}$ is decomposed into the concatenation of two relative positional vectors:
$${\bf p}_{a-i,b-j} \equiv ({\bf p}^x_{a-i} || {\bf p}^y_{b-j}),$$
where $||$ is the concatenation operator and ${\bf p}^x_{a-i},{\bf p}^y_{b-j} \in \mathbb{R}^{C_{out}/2}$ are the relative positional encoding vectors in the X and Y dimensions, respectively, and they are learned together with the model parameters using standard stochastic gradient descent.

\subsection{Attention-based upsampling}
\label{sec:methods_b}
We propose a flexible attention-based upsampling layer that can be used as a drop-in replacement for the strided transposed convolution layer. Given ${\bf X} \in \mathbb{R}^{C_{in}\times H \times W}$, the output of the attention-based upsampling layer is a set of feature maps: ${\bf Z} \in \mathbb{R}^{C_{out}\times H*S \times W*S}$, where $S$ is the upsampling factor. As in attention-based convolution, we apply three $1\times 1$ convolutional layers on ${\bf X}$ to get the query, key, and value tensors: ${\bf Q},{\bf K},{\bf V} \in \mathbb{R}^{C_{out}\times H \times W}$ (see Eq.~\ref{eq:three_convs}). We zero-upsample ${\bf K}$ and ${\bf V}$ by a factor $S$ (see Eq.~\ref{eq:zero_upsample}) and use bilinear interpolation to upscale ${\bf Q}$ by a factor $S$:
\begin{align}
  {\bf K}_{up} &= zero\_upsample({\bf K};S), \\
  {\bf V}_{up} &= zero\_upsample({\bf V};S), \\
  {\bf Q}_{up} &= bilinear\_interpolate({\bf Q};S), 
\end{align}

to obtain ${\bf Q}_{up},{\bf K}_{up},{\bf V}_{up} \in \mathbb{R}^{C_{out}\times H*S \times W*S}$. We define an attention mask ${\bf M}\in \mathbb{R}^{H*S \times W*S}$:
\begin{equation}
  M[i,j] = \begin{cases}
    0 & \parbox{3cm}{if ($i$ mod $S=0$) and ($j$ mod $S=0$)} \\
    -\inf, & \text{otherwise},
  \end{cases}
  \label{eq:mask}
\end{equation}
The mask ${\bf M}$ thus has the value $-inf$ at all the zero positions of the zero-upsampled tensors. We then use a masked attention mechanism to obtain the output feature maps:

\begin{align}
    {\bf z}_{ij} = \sum\limits_{a,b \in \mathcal{N}_K(i,j)}softmax_{ab}({\bf q}_{ij}^T({\bf k}_{ab} & + {\bf p}_{a-i,b-j}  \nonumber \\
    & + M[a,b]){\bf v}_{ab},\label{eq:masked_attn}
\end{align}
where we use the same notation and positional encoding scheme as Eq.~\ref{eq:vanilla_attn_conv}. Wherever the mask value is $-inf$, it effectively sets the attention coefficients to zero at these positions due to the action of the softmax. In effect, this causes these positions to be ignored by the attention mechanism. The attention-based upsampling mechanism is illustrated in Fig.~\ref{fig:attention_upsampling}. By using bilinear interpolation, we generate query vectors at the new spatial positions. The attention mechanism then uses the high-resolution query map to aggregate information only from the valid (non-zero) positions in the zero-upsampled values map using the key vectors at these valid (non-zero) positions.


Our proposed upsampling layer can also go beyond the strided transposed convolution layer by allowing the attention-based fusion of features from different modalities. In joint image upsampling tasks, for example, the goal is to upsample a low-resolution image of one modality (the target modality) using a high-resolution image from a different modality (the guide modality). In this setting, we thus have two sets of input features maps: the guide feature maps ${\bf X}^{HR}\in \mathbb{R}^{C_{hr} \times (S*H) \times (S*W)}$, and the target feature maps ${\bf X}^{LR}\in \mathbb{R}^{C_{lr} \times H \times W}$ that we want to upsample by a factor $S$. In this setting, we generate the query and key tensors from the high resolution guide maps, and the value tensor from the low resolution target maps:
\begin{align}
  {\bf Q}_{up} & = conv1d({\bf X}^{HR},{\bf W}^Q), \\
  {\bf K}_{up} & = conv1d({\bf X}^{HR},{\bf W}^K), \\
  {\bf V}_{up} & = zero\_upsample(conv1d({\bf X}^{LR},{\bf W}^V);S)
  \label{eq:joint_usampling}
\end{align}
We then use masked attention (Eqs.~\ref{eq:mask} and~\ref{eq:masked_attn}) to obtain the output feature maps from the query, key, and value tensors. 

\begin{figure*}[!t]
\centering
\includegraphics[width=0.9\textwidth]{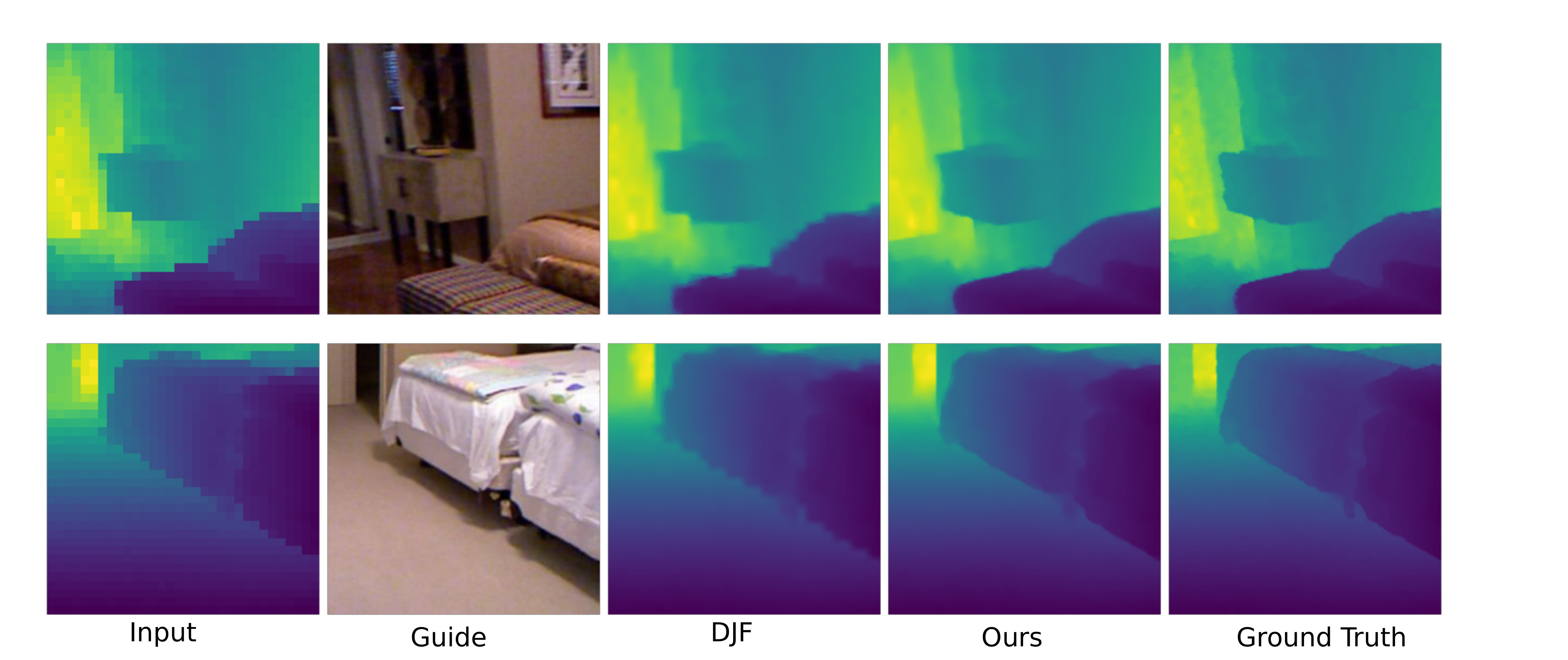}
\caption{Performance of our attention based joint upsampling method as compared to Deep Joint Filtering~\cite{li2016deep} on two target-guide image pairs. For this visual comparison task we use an upsampling factor of $8\times$. Our attention-based method typically produces sharper edges that are in-line with the edges in the guide image while DJF has trouble transferring the high-resolution edge information from the guide image to the upsampled target image.}
\label{fig:djf_localsa_compare}
\end{figure*}

The number of parameters in a traditional strided transposed convolution operation is $C_{in}C_{out}K^2$ (ignoring biases), whereas the number of parameters in our attention-based upsampling operation is $3C_{in}C_{out} + KC_{out}$; these are the parameters of the 3 $1\times1$ convolution layers used to generate the key, query, and value tensors, and the positional encoding parameters. Our attention-based upsampling operation has significantly fewer parameters: approximately ~3X less parameters when $K=3$ and the parameter advantage grows as the kernel size $K$ increases. We will show in the single-image super-resolution tasks that using our attention-based upsampling mechanism as a drop-in replacement of strided transposed convolution not only lowers the parameter count significantly, but also improves accuracy.  


\begin{figure*}[t!]
    \centering
        \subfloat[\label{subfig:plot_1}]{\includegraphics[width=0.25\textwidth]{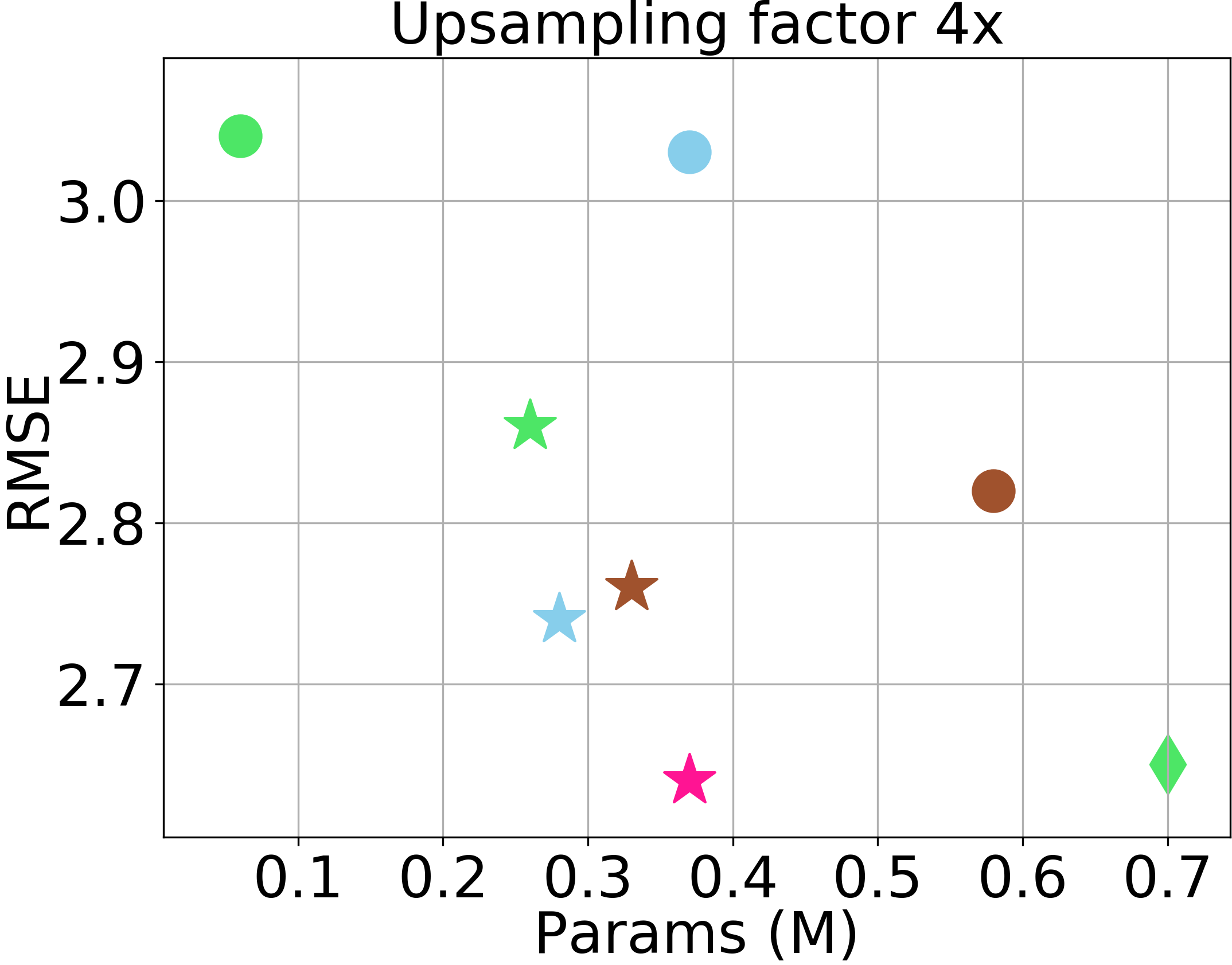}}\hspace*{0.5cm}
        \subfloat[\label{subfig:plot_2}]{\includegraphics[width=0.25\textwidth]{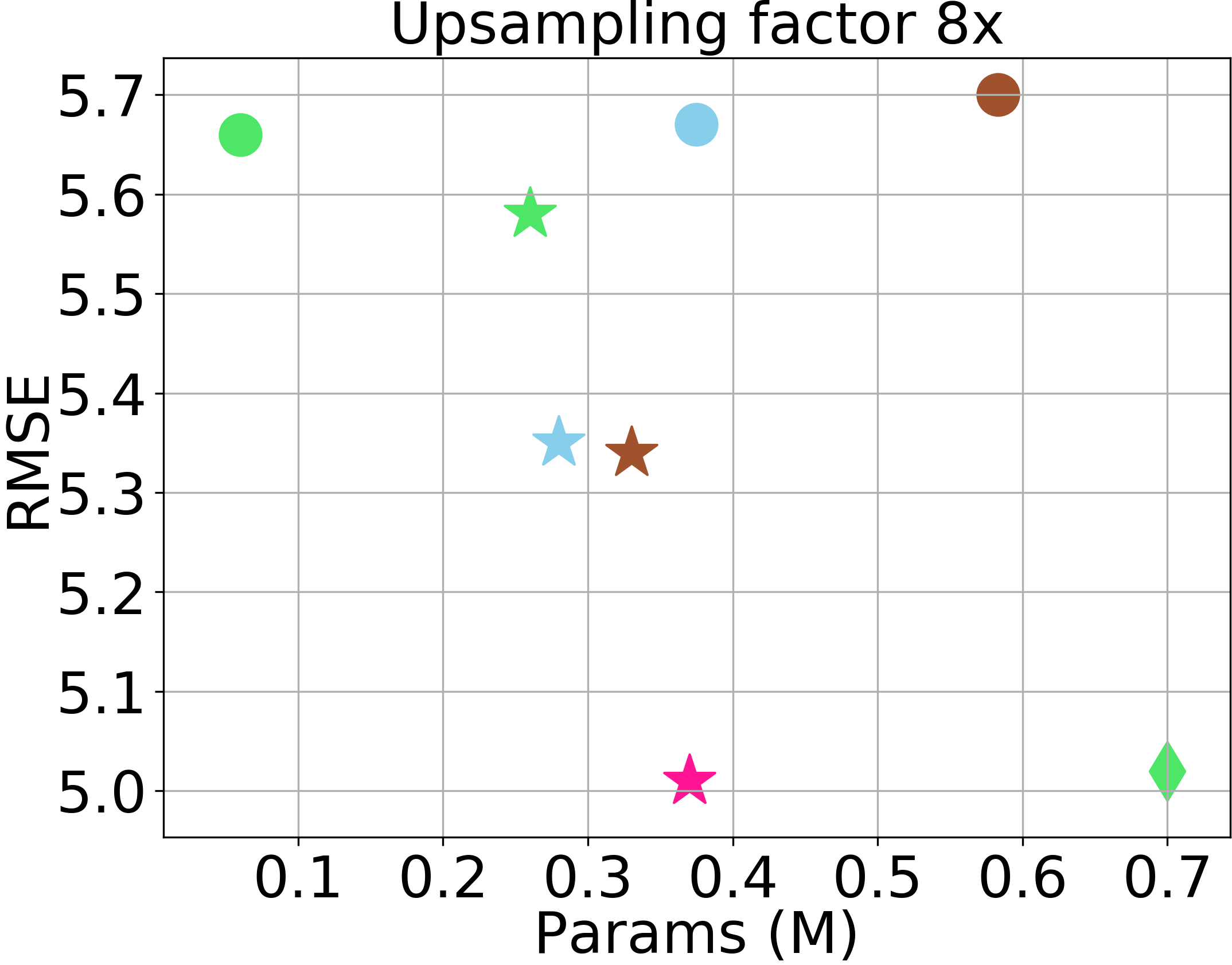}}\hspace*{0.5cm}
        \subfloat[\label{subfig:plot_3}]{\includegraphics[width=0.38\textwidth]{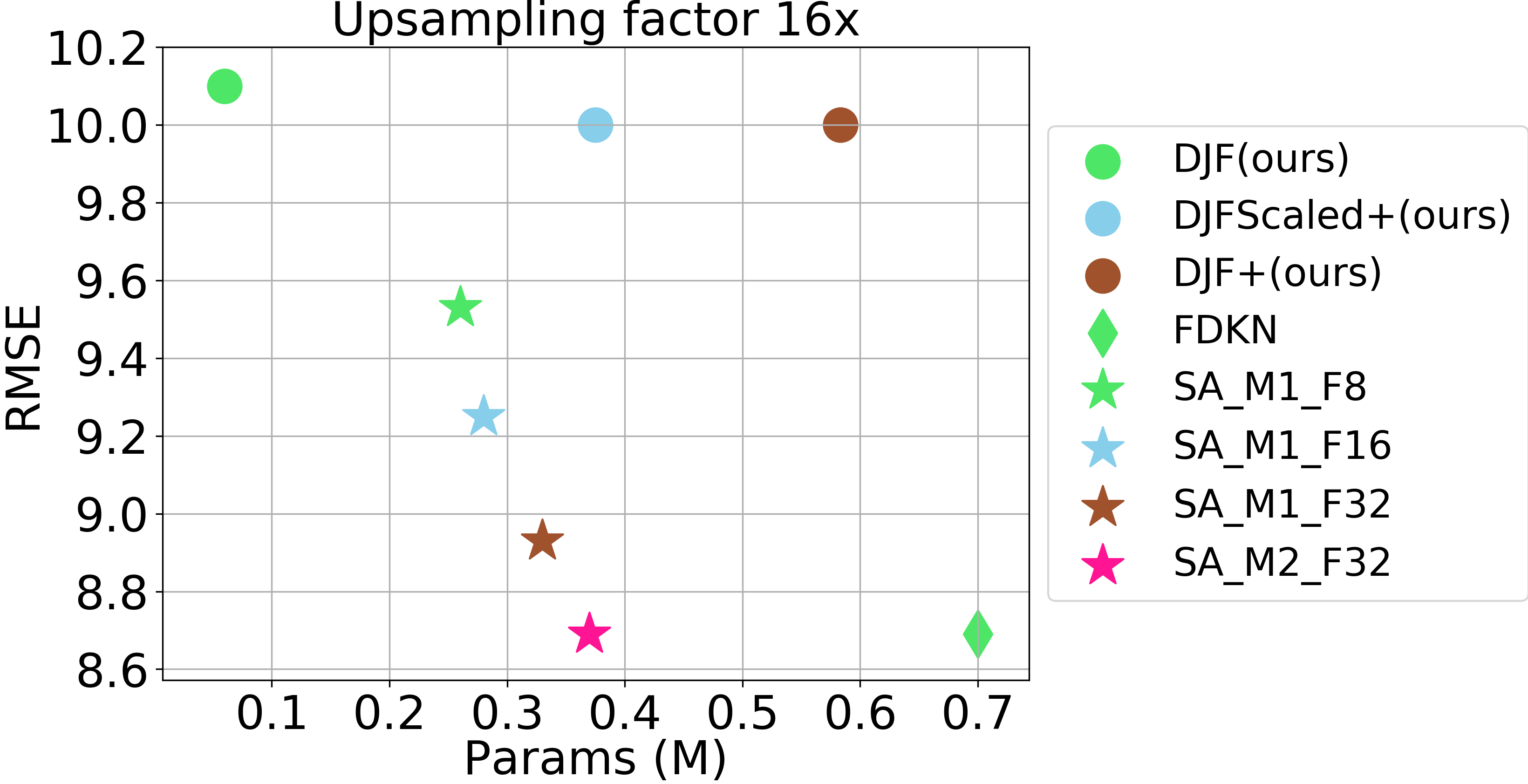}}\vfill
    \caption[Optional caption for list of figures]{Parameters vs RMSE plot for the convolution based DJF, FDKN and our proposed self-attention based models for upsampling factor \subref{subfig:plot_1} $4\times$, \subref{subfig:plot_2} $8\times$, and \subref{subfig:plot_3} $16\times$. It is clear that, for all three upsampling factors, our proposed attention-based upsampling models result in better parameter-RMSE trade-offs.}
    \label{fig:param_vs_rmse}
\end{figure*}

\section{Experimental Analysis}
\label{sec:expt_res}

\subsection{Joint Image Upsampling}
\subsubsection{Architectural Details}

The model architecture is shown in Fig.~\ref{fig:djf_localsa_block} where we use upsampling factor of 4 as an example. The general approach to design the architecture is described in algorithm~\ref{alg:joint_upsampling}. The low-resolution target image and the high-resolution guide image are each first passed through a 3-layer CNN to produce a low resolution set and a high resolution set of feature maps: ${\bf Y}^{LR}$ and ${\bf Y}^{HR}$, respectively.  We then successively upsample by a factor of 2 across multiple attention-based upsampling layers (we thus require the upsampling factor to be a power of 2). In each upsampling stage, the value tensor is the output of the previous stage, or the zero-upsampled low resolution feature maps for the first stage.The query and key tensors are obtained by applying a CNN ($CNN_M$ in Fig.~\ref{fig:djf_localsa_block}) to a concatenation of  the value tensor and an appropriately downsampled version of the high resolution feature maps ${\bf Y}^{HR}$. The output feature maps of $CNN_M$ are split into two parts to obtain the query and key tensors. For an upsampling factor $2^k$, there are $k$ upsampling stages with each stage upsampling by a factor of 2X. 

In the proposed architecture, the high-resolution guide is used to obtain the key and query tensors for the attention-based upsampling layer. These two tensors define how similar two positions are in the guide modality. They provide a high-resolution source of similarity information to the attention-based upsampling layer. Each position in the upsampled output of the target modality would thus be produced by aggregating information from input positions that are most similar to it in the guide modality. 

\begin{algorithm}
  \caption{Attention-based joint upsampling}
  \label{alg:joint_upsampling}
  \begin{algorithmic}
    \Require: ${\bf X}^{LR}\in \mathbb{R}^{1\times (H/M) \times (W/M)}$, ${\bf X}^{HR}\in \mathbb{R}^{3\times H \times W}$
    \Require: Upsampling factor $M$
    \State ${\bf Y}^{LR} \gets CNN_T({\bf X}^{LR})$
    \State ${\bf Y}^{HR} \gets CNN_G({\bf X}^{HR})$
    \State ${\bf V}_{up}^1 \gets zero\_upsample({\bf Y}^{LR};2)$
    \State $n_{steps} \gets log_2(M)$
    \For{$i \gets 1$ to $n_{steps}$}
    \State ${\bf Y}^{HR}_{DS} \gets downsample({\bf Y}^{HR};M//2^i)$
    \State ${\bf Q}_{up}^i,{\bf K}_{up}^i \gets CNN_M({\bf V}_{up}^i || {\bf Y}^{HR}_{DS} )$
    \State ${\bf V}_{up}^{i+1} \gets self\_attention({\bf Q}_{up}^i,{\bf K}_{up}^i,{\bf V}_{up}^i)$
    \EndFor
    \State ${\bf out} \gets CNN_F({\bf V}_{up}^{n_{steps}+1})$

\end{algorithmic}
\end{algorithm}

\subsubsection{Experimental Setup}
We use the NYU depth dataset v2 \cite{Silberman_etal12} which has 1449 pairs of aligned RGB and depth images. We follow \cite{li2016deep}, where we use the first 1000 image-pairs for training and the rest for testing.  We use the high-resolution RGB images to guide and upsample the low-resolution depth map and experiment with upsampling factors of $4\times$, $8\times$, and $16\times$. For each upsampling factor, we obtain the low-resolution depth map by sampling the depth map at grid points that are 4, 8, and 16 pixels apart, respectively. For all these upsampling factors, our goal is to minimize the Root Mean Square Error (RMSE) between the upsampled depth map and the ground truth high resolution depth map. 

We train our network for 2000 epochs with the Adam optimizer and a starting learning rate of 0.001 with a batch-size of 10. We use $256 \times 256$ image crops \footnote{We reduced the learning rate proportionally while using smaller training batches due to memory constraints when using larger models.}. The learning rate is decayed by a factor of 10 after 1200 and 1600 epochs. With this setting, the Deep Joint Filtering(DJF) model~\cite{li2016deep} gives better RMSE than in the original paper. 

\subsubsection{Results}

We use multiple variants of the attention based upsampling model along with increased parameter variants of the original DJF model for upsampling, as described in Table \ref{tab:hybrid_local_sa_models}. In particular, we use different convolutional layers in the mixing block and adjust the channel size to control the parameter count of the proposed model. For example, the 'SA\_M2\_F32' means the self-attention based upsampling model will have 2 convolutional layers in the $CNN_M$ with $2F=64$ channels (refer to Fig. \ref{fig:djf_localsa_block}). For the Fast deformable Kernel Network (FDKN) \cite{Kim_etal20} model we use the  results reported in the original paper. 

\begin{table}[h!]
\begin{center}
\begin{tabular}{|c|c|c|c|}
\hline
Model Name  &\multicolumn{3}{|c|}{Details of CONV layer channel widths} \\
\cline{2-4}
    & $CNN_{T/G}$ & $CNN_F$ & $CNN_M$ \\
\hline
\hline
DJFScaled+ & [384, 192, 1] & [288,144] & -- \\
\hline
DJF+ & [512, 256, 1] & [384,192] & -- \\
\hline
SA\_M1\_F8 & [96, 48, 8] & [16,16] & [16] \\
\hline
SA\_M1\_F16 & [96, 48, 16] & [32,32] & [32] \\
\hline
SA\_M1\_F32 & [96, 48, 32] & [64,64] & [64] \\
\hline
SA\_M2\_F32 & [96, 48, 32] & [64,64] & [64, 64] \\
\hline
\end{tabular}
\end{center}
\caption{ Descriptions of the various models used for joint image upsampling. For the different component CNNs used in the models, we list the number of channels in the CNN convolutional layers.}
\label{tab:hybrid_local_sa_models}
\end{table}

Table \ref{tab:djf_comparison} compares the performance (RMSE) of our proposed model against prior models. As we can see in this table, our attention-based joint-upsampling model outperforms other approaches for all three upscale factors. 

Fig. \ref{fig:djf_localsa_compare} shows a visual comparison of the upsampled depth maps ($8 \times$). Our model's superior ability to extract common structures from the color and depth images is clearly visible. Compared to DJF, our results show much sharper depth transition without the texture-copying artifacts and thus have closer resemblance to the ground truth.

Fig. \ref{fig:param_vs_rmse} shows how RMSE varies with the number of parameters for various learnable filter based techniques including our approach. Interestingly, the original DJF model saturates in terms of providing better performance with more parameters and thus cannot provide better RMSE with the DJF+ and DJFScaled+ models (the increased parameter variants of DJF). However, this is not true for our proposed attention-based joint upsampling model. As we can see in Fig. \ref{fig:param_vs_rmse}, with the increased parameter count our model performance consistently improves.
Also, the FDKN model, having around $1.9\times$ more parameters under-performs compared to our 'SA\_M2\_F32' model for all the three upsampling factors. In this paper, we focused on improving the performance of our attention-based joint upsampling model by simply increasing the model's number of channels  and the number of convolution layers in $CNN_M$. We believe, however, that better and more parameter efficient models can be obtained by replacing the standard convolutions by attention-based convolutions. This is a topic for future work, however.

\begin{table}[t!]
\begin{center}
\begin{tabular}{|c|c|c|c|}
\hline
Method     & \multicolumn{3}{|c|}{Scale factor} \\
\cline{2-4}
           & $4\times$    & $8\times$      & $16\times$ \\
\hline
\hline
Bicubic    & {8.16}&  {14.22}& {22.32}\\ 
\hline
MRF        & {7.84}& {13.98}& {22.20}\\
\hline
GF \cite{he2012guided} & {7.32}& {13.62}& {22.03}\\
\hline
JBU \cite{kopf2007joint}& {4.07}& {8.29} & {13.35}\\
\hline
Ham {et al.} \cite{ham2015robust}& {5.27}& {12.31}& {19.24}\\
\hline
DMSG \cite{hui2016depth} & {3.78}& {6.37}& {11.16}\\
\hline
FBS \cite{barron2016fast}  & {4.29}& {8.94}& {14.59}\\
\hline
DJF \cite{li2016deep}  & {3.54}& {6.20}& {10.21}\\
\hline
FDKN \cite{Kim_etal20}  & 2.65 & 5.02 & 8.69\\
\hline
Ours (Self-Attention) & \textbf{2.64}& \textbf{5.01}& \textbf{8.63}\\
\hline
\end{tabular}
\end{center}
\caption{\textbf{Joint image upsampling}. Results in RMSE (the lower, the better) on NYU v2 dataset show that our upsampling network consistently outperforms traditional techniques including conventional convolution based approach \cite{li2016deep, li2019joint, Kim_etal20} for different upsampling factors.}
\label{tab:djf_comparison}
\end{table}

\begin{figure*}[!t]
\centering
\includegraphics[width=1.0\textwidth]{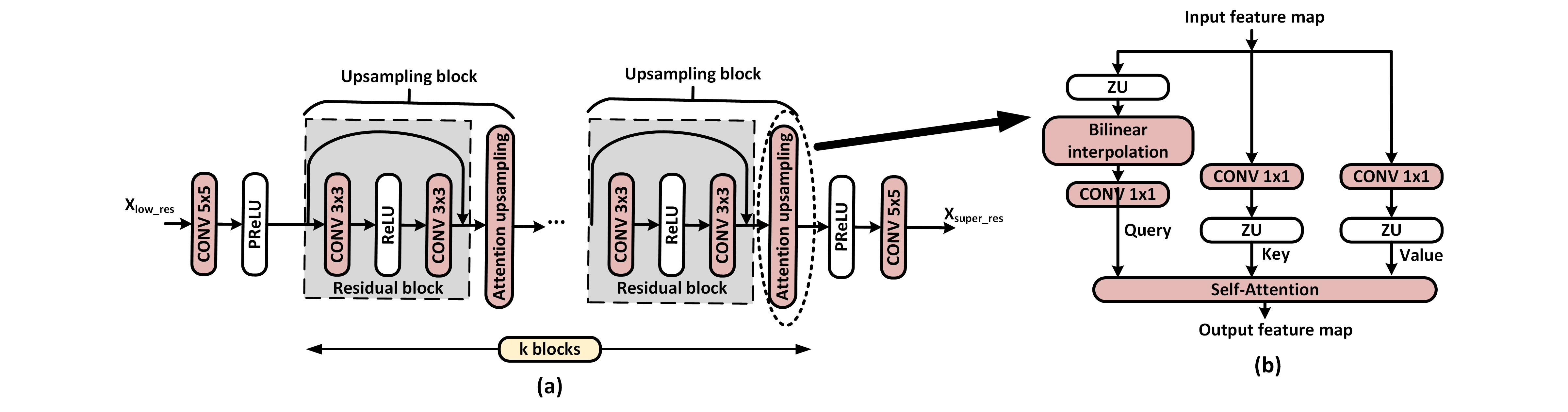}
\caption{(a) Network architecture used for single image super resolution with an upsampling factor of  $2^k$. The proposed model uses an attention-based upsampling block for upscaling the image instead of strided transposed convolution. We use $k$ upsampling blocks (where each upsampling block upsamples the image by a factor of 2X) for an upsampling factor of $2^k$. (b) Internal block diagram of an attention upsampling unit as described in more details in section~\ref{sec:methods_b} and Fig.~\ref{fig:attention_upsampling}. For the baseline model, instead of the attention-based upsampling layer, we use a standard strided transposed convolution layer with stride 2 in each upsampling block.}
\label{fig:sisr_block}
\end{figure*}

\subsection{Single Image Super Resolution}
\subsubsection{Architectural Details}
  The low-resolution input image is first passed through a CNN layer with PReLU activation \cite{he2015delving}. This is followed by an upsampling block which mainly has two components. The first is a residual block having two CNN layers with ReLU activation in between. The second component is the proposed attention-based upsampling layer. For the baseline model, we use a simple strided transposed convolution instead of the attention-based upsampling layer. Each upsampling block upsamples by a factor of 2X. The model architecture used for single image super-resolution with an upsampling factor of $2^k$, having $k$ upsampling blocks is shown in Figure \ref{fig:sisr_block}.
  We use upsampling factors of $2$,$4$ and $8$ for our experiments.

\subsubsection{Experimental Setup}
 We mainly train on two different training sets for our experiments. The $91$ image dataset (T-$91$) is popularly used as the training set for super-resolution \cite{5466111}. Additionally \cite{dong2016accelerating} introduced the General-100 dataset that contains $100$ bmp-format images with no compression. The size of these $100$ images ranges from $710$ × $704$  to $131$ × $112$. We combine both T-91 and General-100 to create one of our training sets. Our second training set is the BSDS-200 \cite{937655} dataset. Additionally, we also perform  data augmentation on the training set similar to \cite{dong2016accelerating}. For augmentation, we downscale images with factors of $0.9$, $0.8$, $0.7$ and $0.6$. Images are additionally rotated at $90$, $180$ and $270$ degrees. To prepare the training samples after augmentation, we initially downsample the original training images by the desired scaling factor to generate the low-resolution (LR) images. The LR images are then cropped into a set of $m \times m$ patches with a fixed stride. The corresponding high-resolution (HR) patches 
are also cropped from the ground truth images.These LR/HR image patches form the basis of our training set similar to \cite{dong2016accelerating}. For evaluation, we use Set5 \cite{bevilacqua2012low}, Set14 \cite{zeyde2010single} and BSDS100 \cite{937655} datasets. 

We train all networks with a learning rate of $0.001$ , batch size of $20$ with the Adam optimizer.  Additionally we use a learning rate scheduler that decays the learning rate by a factor of $0.8$, each time there is a plateau in the evaluation loss. Over the course of training,  we saved the model weights which provided the best PSNR on the evaluation dataset for each run. 

\subsubsection{Results}

Tables \ref{tab:single_img_super1} and \ref{tab:single_img_super2} show the PSNR values(measured in dB) for the deconvolution and attention-based upsampling variants across different evaluation datasets. For statistical significance, the PSNR values shown in Tables \ref{tab:single_img_super1} and \ref{tab:single_img_super2} are an average across three runs with different random seeds. The results for BSDS100, Set5 and Set14 indicate that our proposed architecture performs better than the equivalent deconvolution variant in almost all cases despite having roughly 2x fewer parameters.We hypothesize that this is due to the attention mechanism's context aware feature extraction. Table \ref{tab:params_superes} shows the number of parameters across different scales for both models. This  indicates that self-attention can act as a drop in replacement for strided transposed convolution with much fewer parameters. 

\begin{figure}
	\centering
	\footnotesize
	\vspace{-2mm}
	\begin{tabular}{ccc}
		\hspace{-2mm}
		\includegraphics[width=0.27\linewidth]{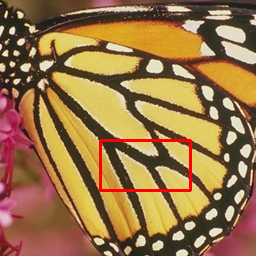} 
		& \hspace{-3mm}
		\includegraphics[width=0.27\linewidth]{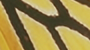}
		& \hspace{-3mm}
		\includegraphics[width=0.27\linewidth]{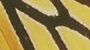}
		\\
		\hspace{-2mm}
		Scale 
		& \hspace{-3mm} 
		2x
		& \hspace{-3mm} 
		2x
		\\
		\\
		\hspace{-2mm}
		\includegraphics[width=0.27\linewidth]{Figs/super-res-results/butterfly/gt_butterfly_overlay.png} 
		& \hspace{-3mm}
		\includegraphics[width=0.27\linewidth]{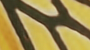} 
		& \hspace{-3mm}
		\includegraphics[width=0.27\linewidth]{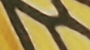} 
		\\
		\hspace{-2mm}
		Scale 
		& \hspace{-3mm} 
		4x
		& \hspace{-3mm} 
		4x
        \\
		\\
		\hspace{-2mm}
		\includegraphics[width=0.27\linewidth]{Figs/super-res-results/butterfly/gt_butterfly_overlay.png} 
		& \hspace{-3mm}
		\includegraphics[width=0.27\linewidth]{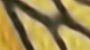} 
		& \hspace{-3mm}
		\includegraphics[width=0.27\linewidth]{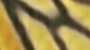} 
		\\
		\hspace{-2mm}
		Scale 
		& \hspace{-3mm} 
		8x
		& \hspace{-3mm} 
		8x
		\\
		\\
		\hspace{-2mm}
		\includegraphics[width=0.27\linewidth]{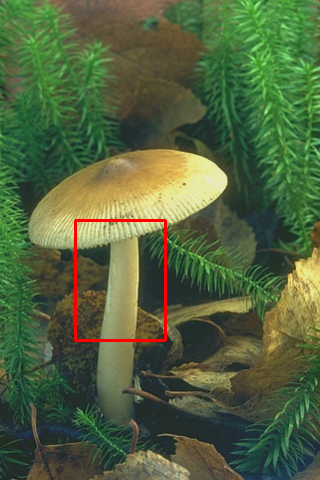} 
		& \hspace{-3mm}
		\includegraphics[width=0.27\linewidth]{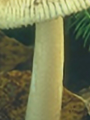}
		& \hspace{-3mm}
		\includegraphics[width=0.27\linewidth]{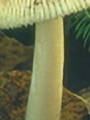}
		\\
		\hspace{-2mm}
		Scale 
		& \hspace{-3mm} 
		2x
		& \hspace{-3mm} 
		2x
		\\
		\\
		\hspace{-2mm}
		\includegraphics[width=0.27\linewidth]{Figs/super-res-results/mushroom/gt_mushroom_overlay.png} 
		& \hspace{-3mm}
		\includegraphics[width=0.27\linewidth]{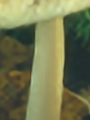} 
		& \hspace{-3mm}
		\includegraphics[width=0.27\linewidth]{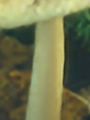} 
		\\
		\hspace{-2mm}
		Scale 
		& \hspace{-3mm} 
		4x
		& \hspace{-3mm} 
		4x
        \\
		\\
		\hspace{-2mm}
		\includegraphics[width=0.27\linewidth]{Figs/super-res-results/mushroom/gt_mushroom_overlay.png} 
		& \hspace{-3mm}
		\includegraphics[width=0.27\linewidth]{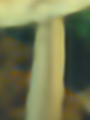} 
		& \hspace{-3mm}
		\includegraphics[width=0.27\linewidth]{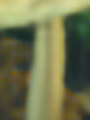} 
		\\
		\hspace{-2mm}
		Scale 
		& \hspace{-3mm} 
		8x
		& \hspace{-3mm} 
		8x
		\\
		\hspace{-2mm}
		(a) Ground Truth HR 
		& \hspace{-3mm} 
		(b) \textbf{Ours}  
		& \hspace{-3mm} 
		(c) Deconvolution
		\\
		\\
	\end{tabular}
	\vspace{-0.3cm}
	\caption{
		\textbf{Single Image Super Resolution}. 
		The first column represents the ground truth image, the second shows the highlighted region in the output of the attention based upsampling model, and the third shows the output of the model using strided transposed convolutions. As the scale increases, the attention based model produces crisper images than the model based on strided transposed convolutions.
	}
	\label{fig:superes-comp}
	\vspace{-2mm}
\end{figure}

Figure \ref{fig:superes-comp} shows the qualitative results for super-resolution across different scales. We used models trained on the combination of T91 \cite{5466111} and General-100 \cite{dong2016accelerating} to generate images shown in Figure \ref{fig:superes-comp}.  We can see that as the upsampling factor increases, our attention-based upsampling model produces crisper images compared to the  strided transposed convolution model. 


\begin{table}[ht!]
\begin{center}
\begin{tabular}{|c|c|c|c|}
\hline
Scale & Attention-based & Deconvolution & Parameter  \\
      & upsampling  &              & Improvement\\       
\hline
\hline
 2 & 0.35 M & 0.71 M & $2.02\times$ \\ 
 4 & 0.69 M & 1.41 M & $2.03\times$ \\ 
 8 & 1.04 M & 2.12 M & $2.03\times$ \\ 
\hline

\end{tabular}
\end{center}
\caption{\textbf{Parameter Count For Single Image Super-resolution}. The total parameter count (in millions) for the proposed attention based upsampling architecture and the deconvolution baseline models. Note that parameters from the CNN layers in the residual blocks and the first CNN layer are also included in the total count. Our model has roughly 2x fewer parameters compared to the baseline.}
\label{tab:params_superes}
\end{table}

\begin{table}[ht!]
\begin{center}
\begin{tabular}{|c|c|c|c|}
\hline
Test dataset & Scale & Attention-based & Deconvolution \\
             &       & upsampling &   \\
\hline
\hline
\multirow{3}{4em}{BSDS100} & 2 & \textbf{33.22} & 33.19 \\ 
& 4 & \textbf{27.84} & 27.76 \\ 
& 8 & \textbf{24.72} & 24.60 \\ 
\hline
\multirow{3}{4em}{Set5} & 2 & \textbf{36.88} & 36.86 \\ 
& 4 & \textbf{30.66} & 30.44 \\ 
& 8 & \textbf{25.56} & 25.42 \\ 
\hline
\multirow{3}{4em}{Set14} & 2 & \textbf{32.52} & \textbf{32.52} \\ 
& 4 & \textbf{27.48} & 27.37 \\ 
& 8 & \textbf{23.87} & 23.81 \\
\hline

\end{tabular}
\end{center}
\caption{\textbf{Single Image Super-resolution}. Results in PSNR measured in $dB$ (the higher, the better) on the BSDS100, Set5 and Set14 datasets show that our attention-based upsampling network consistently outperforms the deconvolution based counterpart for different upsampling factors. The results here showcase the performance of the models trained on a combination of the T-91 \cite{5466111} and General-100 \cite{dong2016accelerating} datasets.}
\label{tab:single_img_super1}
\end{table}

\begin{table}[ht!]
\begin{center}
\begin{tabular}{|c|c|c|c|}
\hline
Test Dataset & Scale & Attention-based & Deconvolution \\
             &       & upsampling &   \\
\hline
\hline
\multirow{3}{4em}{BSDS100} & 2 & \textbf{33.28} & 33.25 \\ 
& 4 & \textbf{27.82} & 27.78 \\ 
& 8 & \textbf{24.80} & 24.74 \\ 
\hline
\multirow{3}{4em}{Set5} & 2 & 36.84 & \textbf{36.86} \\ 
& 4 & \textbf{30.51} & 30.41 \\ 
& 8 & \textbf{25.43} & 25.34 \\ 
\hline
\multirow{3}{4em}{Set14} & 2 & \textbf{32.51} & 32.48 \\ 
& 4 & \textbf{27.45} & 27.37 \\ 
& 8 & \textbf{23.87} & 23.82 \\
\hline

\end{tabular}
\end{center}
\caption{\textbf{Single Image Super-resolution}. Results in PSNR measured in $dB$ (the higher, the better) on the BSDS100, Set5 and Set14 datasets. We observe similar trends as in Table~\ref{tab:single_img_super1} where our attention-based model outperforms the model that uses deconvolution operations. The results here showcase the performance of the models trained on only the BSDS200 dataset \cite{937655}.}
\label{tab:single_img_super2}
\vspace{-3mm}
\end{table}

\section{Conclusion}
\label{sec:concl}
Convolutions are a standard mechanism to learn translation equivariant features. Translation equivariance is achieved by sweeping a fixed and small kernel across the input domain. Untying the kernel weights across spatial positions increases the expressiveness of the network~\cite{Taigman_etal14}. However, parameter count increases dramatically and translation equivariance is lost making this approach suitable only in cases where the objects of interest are aligned in the different images. Kernel generating networks generate convolutional kernels dynamically from the output of auxiliary neural networks~\cite{Jia_etal16,kim2020deformable}, thereby keeping parameter count low while allowing the convolutional kernel to vary across spatial positions. However, due to the use of auxiliary kernel-generating networks, these methods incur additional parameter and computational cost compared to vanilla CNNs. 

Attention-based convolutions allow the kernels to vary across spatial positions while simultaneously reducing parameter count compared to a standard convolution~\cite{Vaswani_etal17,Luong_etal15,Graves_etal14} and maintaining translation equivariance. In this paper, we have shown that attention mechanisms can also be used to replace traditional strided transposed convolutions. The attention-based upsampling mechanism we introduce simultaneously improves accuracy while greatly reducing the parameter count compared to conventional strided transposed convolution in single image super resolution tasks. Moreover, our attention-based upsampling mechanism naturally allows us to integrate information from multiple image modalities. This is done by generating the query and key tensors from one modality, and the value tensor from a different modality. We have shown that this allows us to naturally solve joint-image upsampling tasks and achieve competitive performance. 

Even though our attention-based upsampling mechanism has fewer parameters than standard strided transposed convolution, and only a slightly higher number of floating point operations (FLOPs), training is considerably slower. This is a common issue across convolutional attention mechanisms~\cite{Zhao_etal20,Bello_etal19,Ramachandran_etal19}. The root of the problem is that vanilla convolutions/deconvolutions operations have been heavily optimized across GPU and CPU platforms with several high performance compute kernels available from major hardware vendors and deep learning libraries such as Pytorch and Tensorflow. This is not the case for attention-based convolutions and for our attention-based upsampling mechanisms which lack high quality low-level implementations from hardware vendors or deep learning libraries. Implementing high quality low-level compute kernels for attention-based convolutions/deconvolutions which are efficient across different input dimensions, strides, kernel sizes, etc.. and across different hardware platforms is not an easy task. We thus believe that more widespread support for optimized attention-based convolutions/deconvolutions operations in  popular deep learning libraries is the logical next step to translate the theoretical advantages attention-based convolutions/deconvolutions have in terms of reduced parameter count into practical advantages in terms of reduced hardware power consumption and wall-clock run times.


%
\bibliographystyle{IEEEtran}
\bibliography{./biblio.bib}


\end{document}